\documentstyle[eaclap]{article}

\title{\vspace{-0.5in} Incorporating ``Unconscious Reanalysis'' into
  an  Incremental,  Monotonic Parser}

\author{Patrick Sturt\thanks{
The work reported here was done very much in a collaborative spirit
with my supervisor, Dr. Matthew Crocker, and thanks are due to him for
innumerable suggestions and ideas. I would also like to thank the
people who have offered insightful comments on this work,  in
particular,   David Milward and Martin Pickering.
The research was supported by ESRC grant R00429334338}\\
        Centre for Cognitive Science\\
        Edinburgh\\
        UK\\
        sturt@cogsci.ed.ac.uk\\}
\begin{document}

\maketitle
\vspace{-0.5in}

\begin{abstract}
This paper describes the author's implementation of a parser aimed at
reproducing,  in a computationally explicit system,  the
constraints of a particular psycholinguistic model (Gorrell in
press). In Gorrell's  model, ``unconscious'' garden paths
may be processed via the addition of structural relations to a
monotone increasing set at the point of disambiguation, but there is
no discussion as to {\em how} the parser decides which relations to
add. We model this decision as a search for a node in the tree at
which an explicitly defined parsing operation,  {\em  tree-lowering}
may be applied. With reference to English and Japanese processing
data, we show  the importance of this search for  empirical
adequacy of the  psycholinguistic model.
\end{abstract}

\section{Conscious and Unconscious Garden Paths}

Certain researchers in the psycholinguistic community (Pritchett
(1992), Gorrell (in press)), have argued for a binary distinction
between two distinct types of garden path sentences. {\em Conscious
  garden paths}, such as (1) below,  are locally ambiguous sentences
which give rise to
reanalysis that is both experimentally detectable and causes a
conscious sensation of difficulty or ``surprise effect''.
{\em Unconscious garden paths}, on the other hand, such as (2),  cause
reanalysis which is experimentally detectable,  but
which is generally  not ``noticed'' by the speaker or hearer.

(1) While John was eating the ice cream {\bf melted}.

(2) John knows the truth {\bf hurts}.

This binary distinction has often been used to motivate a two-level
architecture in the human syntactic processing system, where  what
we will call the ``core parser'' performs standard attachment, as well
as being able to reanalyse in the easy cases (such as on reaching {\em
  hurts} in (2)), but where the assistance of a {\em higher level
  resolver}  (to use
Abney's terminology (1987, 1989)), is required to solve the difficult
cases, (such as on reaching {\em melted} in (1)). This ``core parser''
has  been the subject of a number of
computational implementations, including Marcus's deterministic
parser (1980), Description theory (henceforth, D-theory) (Marcus et al
(1983)), and  Abney's
licensing based model (1987, 1989). It has also been the subject of a
number of psycholinguistic studies on a more theoretical level
(Pritchett (1992), Gorrell (in press)).

The implementation described in this paper is based on the most recent
model, that of (Gorrell (in press)). This model is interesting in
that  it does not
allow the parser to employ delay tactics,  such as using a  lookahead
buffer  (Marcus (1980),
Marcus et al (1983)), or  waiting for the head of a phrase to appear
in the input before constructing that phrase (Abney (1987, 1989),
Pritchett  (1992)). Instead, processing is guided by the principle of
{\em Incremental Licensing}, which states that ``the parser attempts
incrementally to satisfy the principles of grammar''. For the purposes
of this implementation, I have interpreted this to  mean that each
word must be attached into a fully-connected  phrase marker as it is
found in the input.\footnote{In fact, Gorrell conjectures that,
  where there is insufficient grammatical information to postulate a
  structural relation between two constituents, such as in a sequence
  of two non-case marked NPs in an English centre-embedded construction, the
  parser may  hold these constituents unstructured in its memory (in
  press, p.212). However, for the purposes of this implementation, we
  have taken the most constrained position. Note that, since we do not
  deal with such constructions, none of the arguments presented here
  hinge on whether or not the parser may buffer material in this way.  }
 The psychological desirability of such a {\em Full
  Attachment} model has been argued for, especially with regard to the
processing of head-final languages, where evidence has been found of
pre-head structuring (Inoue \& Fodor (1991), Frazier
(1987)). Such models  have also been explored  computationally
(Milward (1995), Crocker (1991)).

\section{D-theory and Gorrell's Model}

Gorrell employs the D-theoretic device of building up a set of
dominance and precedence relations\footnote{The original D-theory
  model did  not
  compute precedence relations, except between terminal nodes.}
between  nodes,  where the set
is intended to be constrained by informational monotonicity, in that
once asserted to the set, no relation may be deleted or
overridden. Gorrell restricts this constraint to {\em Primary
  structural relations} (i.e. dominance and precedence), while {\em
  secondary relations} (e.g. thematic and case dependencies)  are not
so constrained.  Recall (2), repeated
below:

(2) John knows the truth hurts.

 At the point  where {\em  John knows the truth} has been processed,
 a complete clause will have been built:

(3)  [$_{S}$ [$_{NP_1}$ John] [$_{VP}$ [$_V$ knows] [$_{NP_2}$ the
truth]]

 The description will include the information that the verb {\em
   knows} precedes {\tt NP$_{2}$}, and that the {\tt VP} dominates {\tt
   NP$_2$}.

\{..., {\tt prec(V,NP$_{2}$), dom(VP, NP$_{2}$)}, ...\}

However,  on the subsequent input of {\em hurts}, the structure can be
reanalysed by asserting an extra clausal node (call it {\tt S$_2$})
dominating  {\tt NP$_2$} (which will then become the embedded subject), but
which is in turn dominated by the matrix VP. This can be achieved by
adding the following structural relations to the tree description
\{prec(V,S$_2$), dom(VP,S$_2$), dom(S$_2$,NP$_2$)\}

(4). [$_{S}$ [$_{NP_1}$ John] [$_{VP}$ [$_V$ knows] [$_{S_2}$ [$_{NP_2}$
the truth] [$_{VP_2}$ hurts]]]]

 Since the description before the processing of the disambiguating word {\em
  hurts} is a subset of the final tree description,
the monotonicity requirement is satisfied. Note in particular, that, because
dominance is a transitive relation, and because of the inheritance
condition on trees (a node inherits the precedence relations of its
ancestors\footnote{See Partee et al (1993) for a description of the
  conditions on
trees, with which all tree descriptions must  comply.}), the two
statements  {\tt dom(VP,NP$_2$)} and {\tt prec(V,NP$_2$)} remain true after
reanalysis.\footnote{It will be noticed that the reanalysis here
  involves a realignment of thematic  and, on GB
  assumptions, case dependencies. These are examples of what Gorrell
  calls {\em secondary relations}, which are not subject to the
  monotonicity requirement.}

Note also that the model will correctly fail to reanalyse for sentence
(1)  above, since the reanalysis will require the retraction of the
domination relation between the VP of the adverbial clause and the NP
{\em the ice cream}.

\section{Implementation}

Although Gorrell proposes a general principle to guide initial
attachment decisions ({\em Simplicity: No vacuous structure
  building}), and specifies the conditions under which ``unconscious
reanalysis'' may occur, the model leaves unspecified the problem  of
how the system may be implemented. Of particular
interest is the problem of how the parser decides {\em which}
relations to add to the set at each point in time, especially at
disambiguating points.

\subsection{Lexical Representation}

The basic framework on which the implementation is built is similar to
Tree Adjoining Grammar (Joshi et al 1975).
  Each lexical category is associated with a set of structural
  relations,  which determine its {\em lexical subtree}.  We call this
  set the {\em subtree projection} of that lexical category.  For
  example,   the subtree projection for verbs in the English grammar is
  as follows, where {\tt Lex} is a variable which will be instantiated to the
actual verb found in the input.

\{{\tt dom(S,NP), dom(S,VP), dom(VP,V),\\
 dom(V,Lex), prec(NP,VP)}\}

Lexical categories are also associated with lists of left and right
attachment sites. In the above case, {\tt NP}, (which will correspond to the
subject of the verb), will be unified with the left attachment
site. If a transitive verb is found in the input, then
the parser consults the verb's argument structure and creates a
new right attachment site for an NP, asserting also that this new NP is
dominated by {\tt VP} and preceded by {\tt V}.

\subsection{Attachment}

Simple attachment can be performed in two
ways, which are defined below, where the term {\em current tree
  description} is intended to denote the the set of structural
relations built up to that point in processing:

Intuitively, left attachment may be thought of in terms of attaching
the current tree description to the left corner of the projection of
the new word, while right attachment corresponds to
attaching the projection of the new word to the right corner of the
current tree description. They are equivalent to Abney's {\em
  Attach-L} and {\em Attach} respectively.

{\sc definition} {\bf Left Attachment:}\\
Let {\em D} be the current tree description, with root node {\em R}.
Let {\em S} be the subtree projection of the new word, whose left-most
attachment site, {\em A} is of identical syntactic category as {\em
  R}. The updated tree description is $S \cup D$, where {\em A} is
unified with {\em R}.

{\sc definition} {\bf Right Attachment:}\\
Let {\em D} be the current tree description, with the  first right
attachment site {\em A}.  Let {\em S} be the subtree projection of the
new word, whose root {\em R} is of identical syntactic category as
{\em A}.  The updated tree description is $S \cup D$, where {\em A} is
unified with {\em R}.

\subsection{Tree Lowering}

It should be clear that, while simple left and right attachment will
suffice for attaching arguments without reanalysis, it will not allow
us to derive the reanalysis required in example (2). For this, we
intuitively require some means of inserting one tree description  inside
another. Schematically, what we require is illustrated below, where
{\tt [1]} is intended to represent the current tree description built
up after {\em John knows the truth} has been parsed, and {\tt [2]} is
intended to represent the subtree description of the new word {\em
  hurts}.
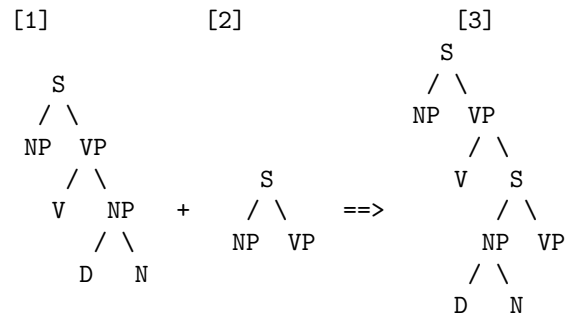
\begin{verbatim}
[1]           [2]               [3]
                               S
   S                          / \
  / \                        NP  VP
 NP  VP                          / \
    / \           S             V   S
   V   NP   +    / \    ==>        / \
      / \       NP  VP            NP  VP
     D   N                       / \
                                D   N

\end{verbatim}

We will call this operation  ``tree-lowering''. Intuitively, the
operation   finds a node on the
current tree description which matches the left attachment site of the
projection of the new word, and attaches it, while inserting the root
of the new projection in its place. The result is that the node chosen
is ``lowered'' or ``subordinated''.

In order to maintain structural coherence, the new word attached via
tree-lowering must be preceded by all other words previously attached
into  the description. We can guarantee this by requiring the lowered node to
dominate the last word to be attached. We also need to ensure that, to
avoid crossing branches, the lowered node does not dominate any
unsaturated attachment sites (or ``dangling nodes'')
 We therefore define {\em   accessibility} for tree-lowering as follows:

{\sc definition }{\bf Accessibility:}\\
Let {\em N} be a node in the current tree description. Let {\em W} be
the last word to be attached into the tree.\\
{\em N} is accessible iff {\em N} dominates {\em W}, and {\em N} does
not dominate any unsaturated attachment sites.

{\sc definition }{\bf Tree-lowering:}\\
Let {\em D} be the current tree description. Let {\em S} be the
subtree projection  of the new
word. The left attachment site {\em A} of {\em S} must match a node
{\em N} accessible in
{\em D}. The root node {\em R} of {\em S}
must be licensed by the grammar in the position occupied by {\em N}.
Let {\em L} be the set of local relations in which {\em N}
participates. Let {\em M} be the result of substituting all instances
of {\em N} in {\em L} with {\em R}. The attachment node {\em A} is
unified  with {\em N}.\\
The updated tree-description is $D \cup S \cup M$\footnote{Note that Abney's
  {\sc steal} operation (1987, 1989) is  more powerful than
  tree-lowering, since it may change domination relations,  and thus
  will allow sentences such as (1), though it excludes reduced relative
  garden paths,  such as
  {\em The horse raced past the barn fell}. The original
  D-theory model (Marcus et al (1983)) is also more powerful, because
  it allows the right-most daughter of a node to be lowered under a
  sibling node. }

It will  be noticed that
tree-lowering is similar in spirit to the adjunction operation of
Tree Adjoining Grammars (Joshi et al, 1975). The difference is that the
   foot and root
nodes of an auxiliary tree in TAG, (corresponding to the ``lowered''
node and the node that replaces it respectively) must be of the same
syntactic
category, whereas, as we have seen in this example, in the model
proposed here, the two nodes may
be of different categories, so long as the resulting structure is
licensed by the grammar.

In the case of example (2), at the point where {\em the truth} has
been processed, the parser must find an accessible node which matches
the category of the left attachment site of {\em hurts} (i.e. an NP).
The only choice is {\tt NP$_2$}:

(3)  [$_{S}$ [$_{NP_1}$ John] [$_{VP}$ [$_V$ knows] [$_{NP_2}$ the truth]]

Now, all the local relations in which {\tt NP$_2$} participates are
found:

\{{\tt dom(VP,NP$_2$), prec(V,NP$_2$)}\}

and {\tt NP$_2$} is substituted with the root of the new projection,
{\tt S$_2$} to derive two new relations:

\{{\tt dom(VP,S$_2$), prec(V,S$_2$)}\}

These relations are found to be licensed, because the verb which {\tt
  V} dominates (``knows'') may subcategorise for a clause, so these
new relations are added to the set\footnote{Note that the relations
defining the original position of {\tt NP$_2$}, (i.e.  {\tt dom(VP,NP$_2$)}
and  {\tt prec(V,NP$_2$)}) are not subtracted from the set. }. Now,
adding the subtree projection of {\em hurts} to the set, and  unifying
its left attachment site with {\tt NP$_2$} results in the derived
structure with {\tt NP$_2$} ``subordinated'' into the lower clause.

 [$_{S}$ [$_{NP_1}$ John] [$_{VP}$ [$_V$ knows] [$_{S_2}$ [$_{NP_2}$ the
 truth] [$_{VP_2}$ hurts]]]]

With the tree-lowering operation so defined, the problem of finding
which relations to add to the set at a disambiguating point reduces to
a search for an accessible node at which to apply this operation.
However, this
implies that, if more than one such node exists, the parser must be
given a preference for making the requisite decision.
Consider the following sentence fragment, for example:

(5) I know [$_{NP_1}$ the man who believes [$_{NP_2}$ the
countess]]...

If the input subsequently continues with a verb, then we have a choice
of two nodes for lowering, i.e. {\tt NP$_1$} and {\tt NP$_2$}. Though
no experimental work has been done on this type of sentence,  there
seems to be an intuitive preference for the lower attachment site, {\tt
  NP$_2$}. In (6), binding constraints force lowering to be applied at
{\tt NP$_2$}, while in (7), it must be applied at {\tt NP$_1$}. Of the
two, most native English speakers report (6) to be easier.

(6) I know  the man who believes  the countess killed herself.

(7) I know the man who believes the countess killed himself.

Note also, that, on standard X-bar assumptions, the attachment of
post-modifiers may be derived via lowering at an X$'$ node. In this
case, the lowered node and its replacement will be of the same
syntactic category (like the root and foot node of a TAG auxiliary
tree). Researchers have noted a general preference for low attachment
of post-modifiers (this is accounted for by the principle of {\em late
  closure} (Frazier and Rayner, 1982)). This would suggest that a
reasonable search strategy for English would be to search the set of
accessible node in a bottom-up direction for English.

The algorithm is constructed in such a way that lowering is only
attempted in cases where simple attachment fails. This means that
arguments (which are incorporated via simple attachment) will be
attached preferentially to adjuncts (which are incorporated via
lowering). This captures the general preference for argument over
adjunct attachment, which is accounted for by the principle of {\em
  Minimal attachment} in Frazier and Rayner (1982), and by the
principle of {\em simplicity} in Gorrell (in press).

\section{Processing Japanese}

\subsection{Main/subordinate clause ambiguity}

Japanese presents a challenge for any incremental parsing model
because, typically, it is not possible to determine where an embedded
clause begins. Consider the following example:

(8)  John ga [\O$_i$ ronbun wo kaita] seito$_i$ wo hometa.\\
 John NOM essay ACC wrote student ACC praised \\
``John praised the student who wrote the essay''

Up to the first verb {\em kaita} (``wrote''), the string is
interpretable as a full clause (without a gap), meaning ``John wrote an
essay'', and the incremental parser builds the requisite structure.
However, the appearance of the head noun {\em seito}
(student) means that at least part of the preceding clause must be
reinterpreted as a relative clause including a gap (note that there is no
 overt relative
pronoun in Japanese).  One way of looking at what is happening here is
to see    the subject NP {\em
  John ga} as being dissociated from the clause in which it is
originally attached, and reattached into  the main clause.
 But looking at it from a different perspective, as Gorrell has noted
 (in press), one can see the subject NP as {\em remaining} in the main
 clause, and the constituent bracketed in (8), ({\em ronbun wo kaita}
 (``wrote an essay''))  as being {\em lowered}
 into the relative clause. If this is possible, then we would expect
 examples like (8) to be unconscious garden paths, and this does
 indeed seem to be reflected in the intuitive data (see Mazuka and
 Itoh (in press)).
However, if we are to allow our parser to handle such examples, we
must expand the definition of tree-lowering, since,
in order to build a relative clause,
we have to assert extra   material (including the empty subject and
the new S node),  which is not justified solely by
the lexical requirements of the disambiguating word, the head noun
{\em seito}. This involves reconstructing all the clausal structure
dominating the lowering site (including asserting empty argument
positions), with reference to the verb's case frame, and attempting to
attach the result as a relative clause to the head noun.

\subsection{Minimal Expulsion}

Inoue (1991), describes a ``minimal expulsion strategy'', which
predicts a preference, on reanalysis, towards expelling the minimum
amount of material from the clause. In our terms, this means that
(assuming a binary right-branching clause structure, with the verb in
its right corner)  the
node selected for lowering must be as high as possible.
This means that the bottom-up search which we
use for English will wrongly predict a {\em Maximal} expulsion
strategy.  In  cases
such as (8), assuming the bottom-up search,  when a post-clausal noun
has been  reached in the input, the parser
starts its search  from the node immediately dominating the last word to
be incorporated, (i.e. the verb of what will become the relative
clause). This means that,  in cases such as (8), the
first preference  will be
to lower the verb (and therefore ``expel'' both  subject and object),
whereas
the human preference, (to lower the object and verb, and therefore
expel only the subject) is the parser's second
choice on the bottom-up search strategy.

Mazuka and Itoh (in press) note that examples where both
subject and object must be expelled from the relative clause, as would
be the first choice in a bottom-up search,
 often cause a conscious
garden path effect. An example, adapted  from Mazuka and Itoh is the
following:

(9) Yamasita ga yuuzin wo [\O{} \O$_i$ houmonsita] kaisya$_i$ de mikaketa. \\
Yamasita NOM friend ACC {} {} visited company LOC  saw

``Yamasita saw his friend at the company he visited.''

In order to capture  the minimal expulsion strategy in this
class  of Japanese examples,
therefore, search  for the lowering
node should be conducted top-down. We are currently investigating the
consequences of changing the search strategy in this way.

\section{The Problem of Retrospective Reanalysis}

Having formulated the constraints of Gorrell's model in terms of the  {\em
  accessibility}  of a node for {\em tree-lowering}, we can see that
the model can be falsified if we can find a  case where the relevant
disambiguating information comes at a point in processing where the
node  which is required  to be lowered   is no longer
accessible. Consider the following pair of sentences:

(10) I saw the man with the moustache.

(11) I saw the man with the telescope.

It is familiar from the psycholinguistic literature that there is a
preference for attaching the {\em with} phrase as an instrumental
argument of the verb (as in (11), on the reading where the telescope
is the instrument of seeing). On the assumption that {\em saw} selects
for a PP instrumental argument, we can derive this preference in the
present model via the preference to attach as an argument as opposed
to an adjunct. However, since we are constrained by incrementality, we
will have to make an  attachment decision for  the PP as soon as the
preposition {\em with} is encountered, and it will be attached in the
preferred reading as a sister of the verb. This means that, in cases
such as  (10), where, on the globally acceptable reading,  the PP is
an  adjunct of the NP {\em the man}, this attachment will have to be
revised, and the PP retrospectively adjoined into the relevant N$'$
node.  However, once the preposition {\em with} has been attached,
the required N$'$ node will no longer be accessible, and a conscious
garden path effect will be predicted, which, intuitively, does not
occur. Note that there is no garden path effect even if the
preposition is  separated from the
disambiguating head noun by a series of adjectives:  (``I saw the man
  with the neat, quaint, old-fashioned moustache/telescope'').

The same result obtains if we abstract away from the particular
implementational details of tree-lowering, and return to the abstract
level at which  Gorrell states his
model. Once the PP has been attached as an argument of the verb, it
can never be reanalysed as the adjunct of the preceding NP, because
the NP will precede the PP before reanalysis, and dominate it after
reanalysis, which is against the  ``exclusivity condition''
on trees  (i.e. no two nodes may stand in both a dominance and a
precedence relation).\footnote{Note that in Marcus et al (1983), since
  precedence relations were not computed for non-terminals, lowering
  into a predecessor was possible, thus (11) would cause no processing
  difficulty. However, presumably, their parser would overgenerate on
  examples such as {\em the horse raced past the barn fell}.}

A similar problem concerns examples such as the following, from Gibson
et al (1993):

(12) the lamps near the paintings of the house [that was damaged in the
flood].

(13) the lamps near the painting of the houses [that was
           damaged in the flood].

(14)  the lamp near the paintings of the houses [that was
          damaged in the flood].

in the above,  Gibson et al have manipulated number agreement to force
low (12), middle (13) and high (14) attachment of the bracketed relative
clause.
The results of their on- and off-line experiments show clearly that
the low attachment (corresponding to 12) is easiest, but the middle
attachment (corresponding to (13)) is most difficult. This behaviour
cannot be captured whether we adopt  a bottom-up or a top-down search
for tree-lowering. However, even if we can  incorporate the required
preferences into the parser, the constraint of incrementality will
force us to make the decision on encountering {\em that}. This means
that, assuming we decide initially to attach low, but number agreement
on {\em was} subsequently forces high attachment, as in (14),  then a
conscious   garden path effect
will be predicted, as lowering cannot derive the reanalysis. This is
true on the abstract level as well, since there will be nodes in the
description which
precede the original low position of the relative clause, but are
dominated by the subsequent high position of the relative
clause. However, intuitively, of the above sentences, it is only (13)
which causes the conscious garden path effect.\footnote{
Preliminary findings suggest that a similar preference rating is employed
in (written) production as well as (reading)
comprehension for these examples.  This can be seen in  Gibson et al's (1994)
 study. This
shows a the {\em LOW} $>$ {\em HIGH} $>$ {\em MID} ordering in the
attachment of the final PP in NPs of the following form found in the
Brown corpus:

 NP$_1$ Prep  NP$_2$ Prep NP$_3$ PP

Of 105 unambiguous PP adjunct attachments, 68\% were low-attached,
26\% high attached and 10\% mid-attached.
However, the question of whether the
  syntactic  structures people   preferentially use
  in production should correspond to the syntactic structures people
  preferentially assign to strings during comprehension is still very
  much an open issue, though see Mitchell and Cuetos (1991) for a view that
  the  experience of previous  input influences parsing decisions.}

\section{Conclusion}

 The current implementation shows that the
success of an abstract model such as Gorrell's depends crucially on
the  computational details
of the processing algorithm used. The search for the lowering site is
of particular importance. In the final section we have seen that the
combination of informational monotonicity with the assumption of strict
incrementality results in a system which is too constrained to capture
all the processing data. Future research will be aimed at determining,
firstly, how we can enrich the  information to which the search
strategy is sensitive in order to provide a better match with human
preferences, and secondly,
which constraints should be relaxed in order to avoid the problem of
undergeneration.

\section*{References}

Abney, S. P. (1987): Licensing and Parsing. {\em Proceedings of NELS}
{\bf 17} p.1-15, University of Massachusetts, Amherst

Abney, S. P. (1989): A computational model of human parsing. {\em
Journal of Psycholinguistic Research} {\bf 18} p.129-144

Crocker, M. W. (1991): {\em A Logical Model of Competence and
Performance in the Human Sentence Processor.} PhD thesis, Dept. of
Artificial Intelligence, University of Edinburgh, Edinburgh, U.K.

Frazier, L. (1987): Syntactic processing: Evidence from Dutch. In {\em
Natural Language and Linguistic Theory} {\bf 5.4} p.519-559

Frazier, L. and K. Rayner, (1982): Making and correcting errors during
sentence comprehension: Eye movements in the analysis of structurally
ambiguous sentences. {\em Cognitive Psychology} {\bf 14} p.178-210

Gibson, E., N. Pearlmutter, E. Canesco-Gonzalez and Greg Hickok
(1993): Cross-linguistic Attachment Preferences: Evidence from English
and Spanish. (ms. submitted to Cognition)

Gibson, E. and  N. Pearlmutter, E. (1994): A corpus-based account of
Psycholinguistic Constraints on Prepositional Phrase Attachment (in
C. Clifton, L. Frazier and K. Rayner (eds) {\em Perspectives on
  Sentence Processing} New York: Lawrence Erlbaum

Gorrell, P. (in press): {\em Syntax and Perception.} to be published
by Cambridge University Press

Inoue, A. (1991): {\em A comparative study of parsing in English and
Japanese.} PhD thesis, University of Conneticut.

Inoue, A. and J.D. Fodor (in press): Information-paced parsing of
Japanese. (to appear in Mazuka \& Nagai (eds))

Joshi, A.K., L.S. Levy, and M. Takahashi, (1975): Tree Adjunct
grammars. {\em Journal of Computer and System Sciences} {\bf 10},
p.136-163

Marcus, M. (1980): {\em A Theory of Syntactic Recognition for Natural
Language} Cambridge, MA: MIT Press

Marcus, M., D. Hindle, and M. Fleck (1983): D-theory: Talking about
talking about trees. {\em Association for Computational Linguistics}
{\bf 21} p.129-136

Mazuka, R. and  K. Itoh  (in press): Can Japanese be led down the
garden path? (to appear in Mazuka and Nagai)

Mazuka, R., and Nagai (eds) (to appear): {\em Japanese Syntactic
Processing} Hillsdale, NJ: Lawrence Earlbaum

Milward, D. (1995): Incremental Interpretation of Categorial
Grammar.  in {\em Proceedings of EACL} (this volume)

Mitchell, D.C. \& Cuetos, F. (1991): The origins of parsing
strategies. {\em Conference proceedings: Current issues in natural
  language processing} University of Texas at Austin, TX

Partee, B., A. ter Meulen and R. E. Wall (1993): {\em Mathematical
  methods in Linguistics} Dordrecht: Kluwer Academic Publishers

Pritchett, B. L. (1992): {\em Grammatical Competence and Parsing
Performance.} Chicago, IL: University of Chicago Press

\end{document}